\documentclass[10pt,twocolumn,letterpaper]{article}

\usepackage{cvpr}
\usepackage{times}
\usepackage{epsfig}
\usepackage{graphicx}
\usepackage{amsmath}
\usepackage{amssymb}
\usepackage{lscape}
\usepackage{enumitem}
\usepackage{multirow}
\usepackage[pagebackref=true,breaklinks=true,letterpaper=true,colorlinks,bookmarks=false]{hyperref}

\cvprfinalcopy 

\def\cvprPaperID{****} 

\ifcvprfinal\fi
\begin{document}

\title{PointFusion: Deep Sensor Fusion for 3D Bounding Box Estimation}

\author{Danfei Xu\thanks{Work done as an intern at Zoox, Inc.}\\
Stanford Unviersity\\
{\tt\small danfei@cs.stanford.edu}
\and
Dragomir Anguelov \\
Zoox Inc.\\
{\tt\small drago@zoox.com}
\and
Ashesh Jain \\
Zoox Inc.\\
{\tt\small ashesh@zoox.com}
}

\maketitle
\newcommand{\vparagraph}[1]{\vspace{-12pt}\paragraph{#1}}

\begin{abstract}
We present PointFusion, a generic 3D object detection method that leverages both image and 3D point cloud information. Unlike existing methods that either use multi-stage pipelines or hold sensor and dataset-specific assumptions, PointFusion is conceptually simple and application-agnostic. The image data and the raw point cloud data are independently processed by a CNN and a PointNet architecture, respectively. The resulting outputs are then combined by a novel fusion network, which predicts multiple 3D box hypotheses and their confidences, using the input 3D points as spatial anchors.  We evaluate PointFusion on two distinctive datasets: the KITTI dataset that features driving scenes captured with a lidar-camera setup, and the SUN-RGBD dataset that captures indoor environments with RGB-D cameras. 
Our model is the first one that is able to perform better or on-par with the state-of-the-art on these diverse datasets without any dataset-specific model tuning.
\end{abstract}

\vspace{-15pt}
\section{Introduction}
We focus on 3D object detection, which is a fundamental computer vision problem impacting most autonomous robotics systems including self-driving cars and drones. The goal of 3D object detection is to recover the 6 DoF pose and the 3D bounding box dimensions for all objects of interest in the scene. While recent advances in convolutional neural networks have enabled accurate 2D detection in complex environments~\cite{ren2015faster,lin2017featpyrnet,lin2017focalloss}, the 3D object detection problem still remains an open challenge. Methods for 3D box regression from a single image, even including recent deep learning methods such as ~\cite{mousavian20163d,subcnn}, still have relatively low accuracy especially in depth estimates at longer ranges. Hence, many current real-world systems either use stereo or augment their sensor stack with lidar and radar. The lidar-radar mixed-sensor setup is particularly popular in self-driving cars and is typically handled by a multi-stage pipeline, which preprocesses each sensor modality separately and then performs a \textit{late fusion} or \textit{decision-level fusion} step using an expert-designed tracking system such as a Kalman filter~\cite{cho2014multi,enzweiler2011multilevel}. Such systems make simplifying assumptions and make decisions in the absence of context from other sensors. Inspired by the successes of deep learning for handling diverse raw sensory input, we propose an \textit{early fusion} model for 3D box estimation, which directly learns to combine image and depth information optimally.

\begin{figure}[!t]
\centering
\includegraphics[width=\linewidth]{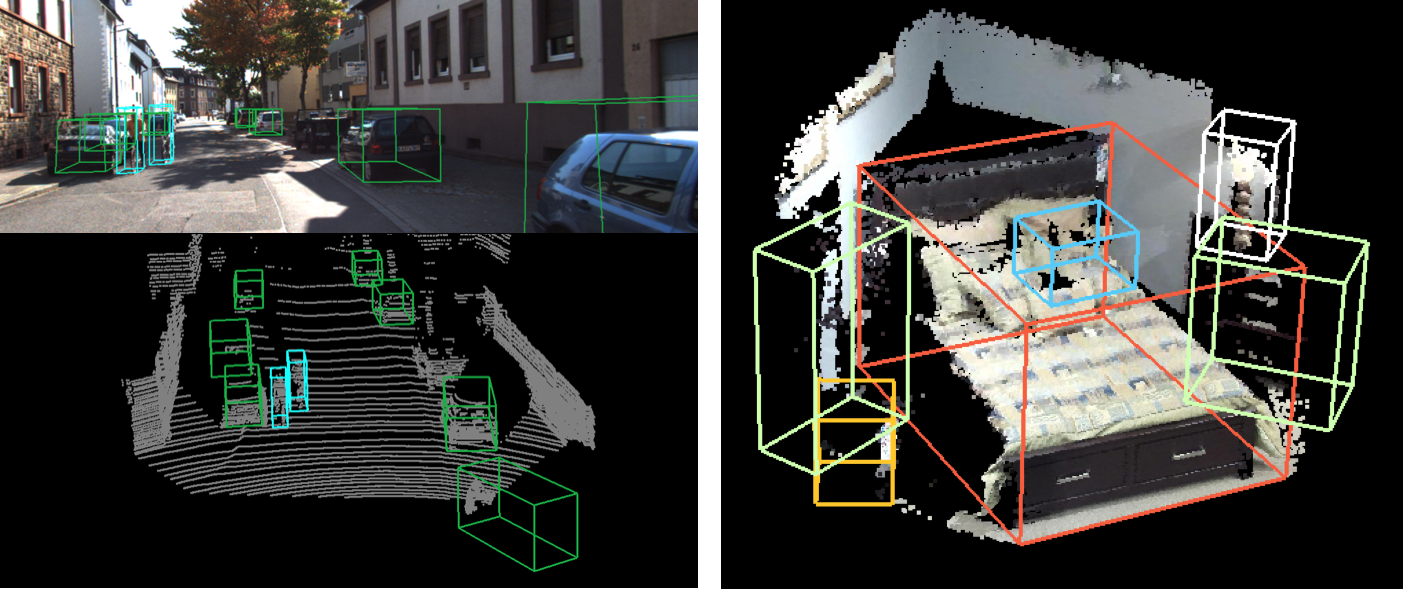}
\caption{Sample 3D object detection results of our ~\textit{PointFusion} model on the KITTI dataset~\cite{kitti} (left) and the SUN-RGBD~\cite{song2015sun} dataset (right). In this paper, we show that our simple and generic sensor fusion method is able to handle datasets with distinctive environments and sensor types and perform better or on-par with state-of-the-art methods on the respective datasets. }
\vspace{-10pt}
\label{fig:pull}
\end{figure}

Various combinations of cameras and 3D sensors are widely used in the field, and it is desirable to have a single algorithm that generalizes to as many different problem settings as possible. Many real-world robotic systems are equipped with multiple 3D sensors: for example, autonomous cars often have multiple lidars and potentially also radars. Yet, current algorithms often assume a single RGB-D camera~\cite{song2016deep,lahoud20172d}, which provides RGB-D images, or a single lidar sensor~\cite{mv3d,velofcn}, which allows the creation of a local \textit{front view} image of the lidar depth and intensity readings. Many existing algorithms also make strong domain-specific assumptions. For example, MV3D~\cite{mv3d} assumes that all objects can be segmented in a  top-down 2D view of the point cloud, which works for the common self-driving case but does not generalize to indoor scenes where objects can be placed on top of each other. Furthermore, the top-down view approach tends to work well for objects such as \textit{cars}, but does not for other key object classes such as \textit{pedestrians} or \textit{cyclists}. Unlike the above approaches,
our architecture is designed to be domain-agnostic and agnostic to the placement, type, and number of 3D sensors. As such, it is generic and can be used for a variety of robotics applications. 

In designing such a generic model, we need to solve the challenge of combining the heterogeneous image and 3D point cloud data. Previous work addresses this challenge by directly transforming the point cloud to a convolution-friendly form. This includes either projecting the point cloud onto the image~\cite{giering2015multi} or voxelizing the point cloud~\cite{song2016deep,li20163d}. Both of these operations involve lossy data quantization and require special models to handle sparsity in the lidar image~\cite{uhrig2017sparsitycnn} or in voxel space~\cite{riegler2017octnet}. Instead, our solution retains the inputs in their native representation and processes them using heterogeneous network architectures. Specifically for the point cloud, we use a variant of the recently proposed PointNet~\cite{qi2016pointnet} architecture, which allows us to process the raw points directly.

\begin{figure*}[t!]
\centering
\includegraphics[width=0.95\linewidth]{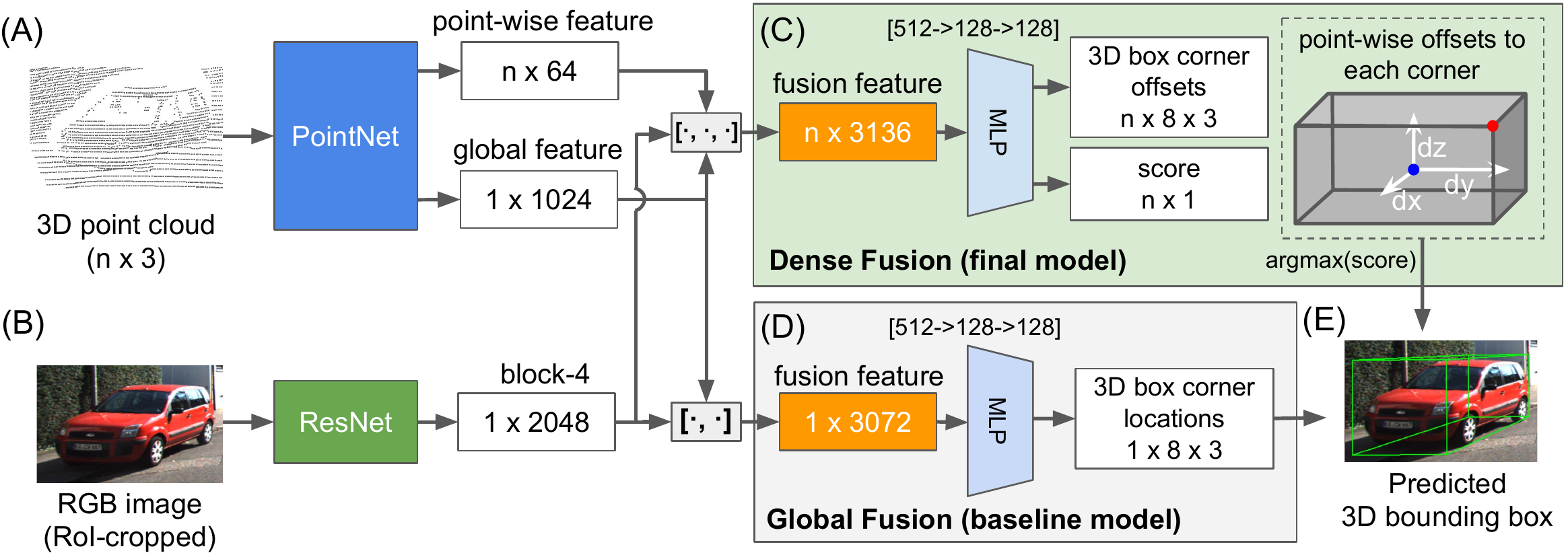}
\caption{An overview of the dense PointFusion architecture. PointFusion has two feature extractors: a PointNet variant that processes raw point cloud data (A), and a CNN that extracts visual features from an input image (B). We present two fusion network formulations: a vanilla \textit{global} architecture that directly regresses the box corner locations (D), and a novel \textit{dense} architecture that predicts the spatial offset of each of the 8 corners relative to an input point, as illustrated in (C): for each input point, the network predicts the spatial offset (white arrows) from a corner (red dot) to the input point (blue), and selects the prediction with the highest score as the final prediction (E).}
\label{fig:model}
\end{figure*}

Our deep network for 3D object box regression from images and sparse point clouds has three main components: an off-the-shelf CNN~\cite{resnet} that extracts appearance and geometry features from input RGB image crops, a variant of PointNet~\cite{qi2016pointnet} that processes the raw 3D point cloud, and a fusion sub-network that combines the two outputs to predict 3D bounding boxes. This heterogeneous network architecture, as shown in Fig.~\ref{fig:model}, takes full advantage of the two data sources without introducing any data processing biases. Our fusion sub-network features a novel \textit{dense} 3D box prediction architecture, in which for each input 3D point, the network predicts the corner locations of a 3D box relative to the point. The network then uses a learned scoring function to select the best prediction. The method is inspired by the concept of spatial anchors~\cite{ren2015faster} and dense prediction~\cite{huang2015densebox}. The intuition is that predicting relative spatial locations using input 3D points as anchors reduces the variance of the regression objective comparing to an architecture that directly regresses the 3D location of each corner. We demonstrate that the dense prediction architecture outperforms the architecture that regresses 3D corner locations directly by a large margin.

We evaluate our model on two distinctive 3D object detection datasets. The KITTI dataset~\cite{kitti} focuses on the outdoor urban driving scenario in which pedestrians, cyclists, and cars are annotated in data acquired with a camera-lidar system. The SUN-RGBD dataset~\cite{song2015sun} is recorded via RGB-D cameras in indoor environments, with more than 700 object categories. We show that by combining PointFusion with an off-the-shelf 2D object detector~\cite{ren2015faster}, we get comparable or better 3D object detections than the state-of-the-art methods designed for KITTI~\cite{mv3d} and SUN-RGBD~\cite{lahoud20172d,song2016deep,ren2016three}. To the best of our knowledge, our model is the first to achieve competitive results on these very different datasets, proving its general applicability.

\section{Related Work}
We give an overview of the previous work on 6-DoF object pose estimation, which is related to our approach. 

\textbf{Geometry-based methods} A number of methods focus on estimating the 6-DoF object pose from a single image or an image sequence. These include keypoint matching between 2D images and their corresponding 3D CAD models~\cite{aubry2014seeing, collet2011moped,zhu2014single}, or aligning 3D-reconstructed models with ground-truth models to recover the object poses~\cite{rothganger20063d,ferrari2006simultaneous}. Gupta~\etal~\cite{gupta2015aligning} propose to predict a semantic segmentation map as well as object pose hypotheses using a CNN and then align the hypotheses with known object CAD models using ICP. These types of methods rely on strong category shape priors or ground-truth object CAD models, which makes them difficult to scale to larger datasets. In contrary, our generic method estimates both the 6-DoF pose and spatial dimensions of an object without object category knowledge or CAD models.

\textbf{3D box regression from images}
The recent advances in deep models have dramatically improved 2D object detection, and some methods propose to extend the objectives with the full 3D object poses. \cite{tulsiani2015viewpoints} uses R-CNN to propose 2D RoI and another network to regress the object poses.  \cite{mousavian20163d} combines a set of deep-learned 3D object parameters and geometric constraints from 2D RoIs to recover the full 3D box. Xiang \etal~\cite{subcnn,xiang2015data} jointly learns a viewpoint-dependent detector and a pose estimator by clustering 3D voxel patterns learned from object models. Although these methods excel at estimating object orientations, localizing the objects in 3D from an image is often handled by imposing geometric constraints~\cite{mousavian20163d} and remains a challenge for lack of direct depth measurements. One of the key contributions of our model is that it learns to effectively combine the complementary image and depth sensor information. 

\textbf{3D box regression from depth data} Newer studies have proposed to directly tackle the 3D object detection problem in discretized 3D spaces. Song \etal~\cite{song2014sliding} learn to classify 3D bounding box proposals generated by a 3D sliding window using synthetically-generated 3D features. A follow-up study~\cite{song2016deep} uses a 3D variant of the Region Proposal Network~\cite{ren2015faster} to generate 3D proposals and uses a 3D ConvNet to process the voxelized point cloud. A similar approach by Li \etal~\cite{li20163d} focuses on detecting vehicles and processes the voxelized input with a 3D fully convolutional network. 
However, these methods are often prohibitively expensive because of the discretized volumetric representation. As an example,~\cite{song2016deep} takes around 20 seconds to process one frame. Other methods, such as VeloFCN~\cite{velofcn}, focus on a single lidar setup and form a dense depth and intensity image, which is processed with a single 2D CNN. Unlike these methods, we adopt the recently proposed PointNet~\cite{qi2016pointnet} to process the raw point cloud. The setup can accommodate multiple depth sensors, and the time complexity scales linearly with the number of range measurements irrespective of the spatial extent of the 3D scene.

\textbf{2D-3D fusion} Our paper is most related to 
recent methods that fuse image and lidar data.  MV3D by Chen \etal~\cite{mv3d} generates object detection proposals in the top-down lidar view and projects them to the front-lidar and image views, fusing all the corresponding features to do oriented box regression. This approach assumes a single-lidar setup and bakes in the restrictive assumption that all objects are on the same spatial plane and can be localized solely from a top-down view of the point cloud, which works for cars but not pedestrians and bicyclists. In contrast, our method has no scene or object-specific limitations, as well as no limitations on the kind and number of depth sensors used.

\section{PointFusion}
\label{sec:model}

In this section, we describe our PointFusion model, which performs 3D bounding box regression from a 2D image crop and a corresponding 3D point cloud that is typically produced by lidar sensors (see Fig.~\ref{fig:pull}). When our model is combined with a state of the art 2D object detector supplying the 2D object crops, such as~\cite{ren2015faster}, we get a complete 3D object detection system. We leave the theoretically straightforward end-to-end model to future work since we already get state of the art results with this simpler two-stage setup. In addition, the current setup allows us to plug in any state-of-the-art detector without modifying the fusion network.

PointFusion has three main components: a variant of the PointNet network that extracts point cloud features (Fig.~\ref{fig:model}A), a CNN that extracts image appearance features (Fig.~\ref{fig:model}B), and a fusion network that combines both and outputs a 3D bounding box for the object in the crop. We describe two variants of the fusion network: a vanilla \textit{global} architecture (Fig.~\ref{fig:model}C) and a novel \textit{dense} fusion network (Fig.~\ref{fig:model}D), in which we use a dense spatial anchor mechanism to improve the 3D box predictions and two scoring functions to select the best predictions. Below, we go into the details of our point cloud and fusion sub-components. 

\subsection{Point Cloud Network}
We process the input point clouds using a variant of the PointNet architecture by Qi \etal~\cite{qi2016pointnet}. 
PointNet pioneered the use of a symmetric function (max-pooling) to achieve permutation invariance in the processing of unordered 3D point cloud sets. The model ingests raw point clouds and learns a spatial encoding of each point and also an aggregated global point cloud feature. These features are then used for classification and semantic segmentation.

PointNet has many desirable properties: it processes the raw points directly without lossy operations like voxelization or projection, and it scales linearly with the number of input points. However, the original PointNet formulation cannot be used for 3D regression out of the box. Here we describe two important changes we made to PointNet.

\vparagraph{No BatchNorm} Batch normalization has become indispensable in modern neural architecture design as it effectively reduces the covariance shift in the input features. In the original PointNet implementation, all fully connected layers are followed by a batch normalization layer. However, we found that batch normalization hampers the 3D bounding box estimation performance. Batch normalization aims to eliminate the scale and bias in its input data, but for the task of 3D regression, the absolute numerical values of the point locations are helpful. Therefore, our PointNet variant has \textit{all} batch normalization layers removed.

\begin{figure}[!t]
\centering
\includegraphics[width=0.6\linewidth]{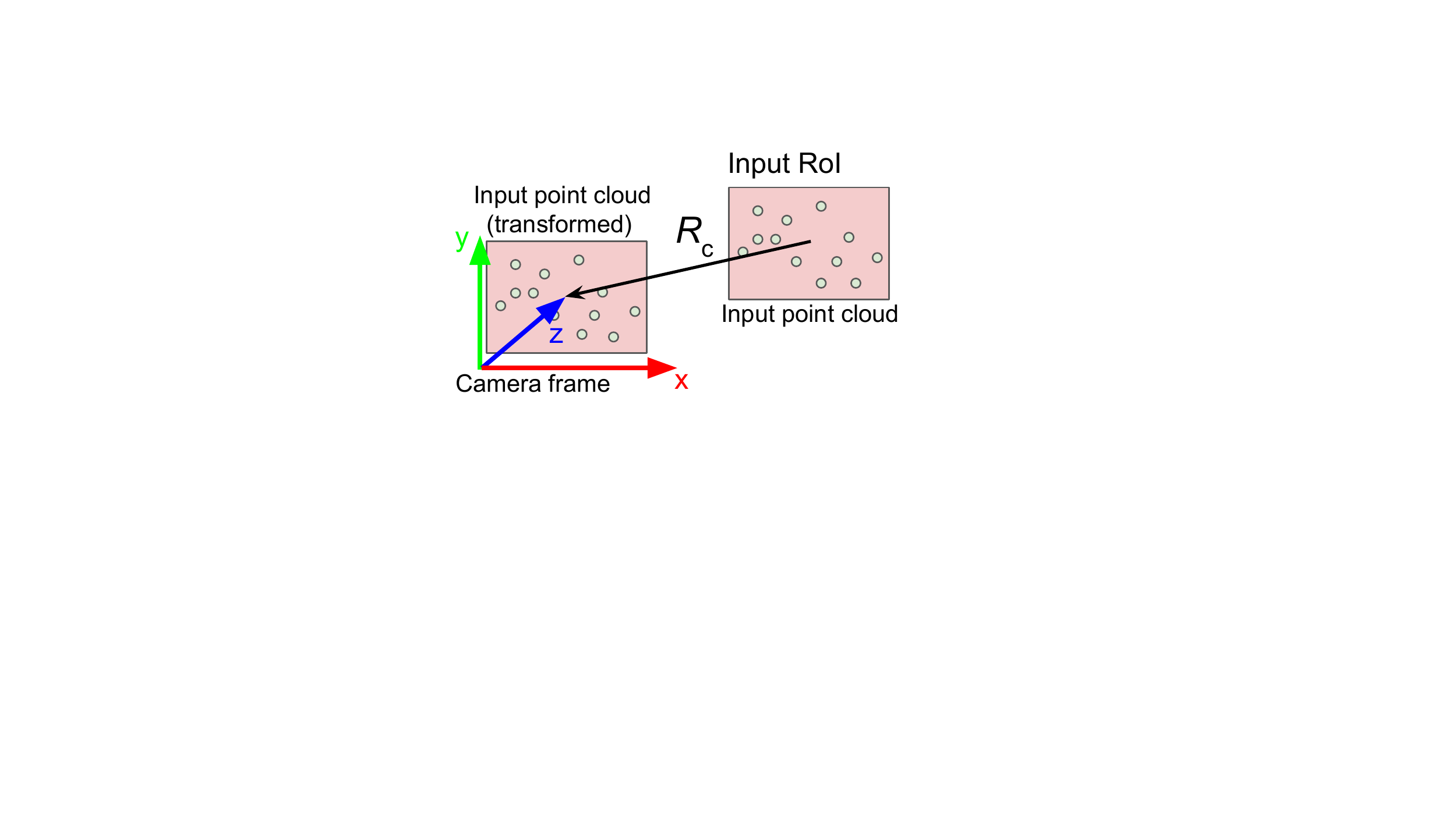}
\caption{During input preprocessing, we compute a rotation $R_c$ to canonicalize the point cloud inside each RoI.}
\label{fig:transform}
\end{figure}

\vparagraph{Input normalization} As described in the setup, the corresponding 3D point cloud of an image bounding box is obtained by finding all points in the scene that can be projected onto the box. However, the spatial location of the 3D points is highly correlated with the 2D box location, which introduces undesirable biases. PointNet applies a Spatial Transformer Network (STN) to canonicalize the input space. However, we found that the STN is not able to fully correct these biases. We instead use the known camera geometry to compute the canonical rotation matrix $R_c$. $R_c$ rotates the ray passing through the center of the 2D box to the $z$-axis of the camera frame. This is illustrated in Fig.~\ref{fig:transform}.

\subsection{Fusion Network}
The fusion network takes as input an image feature extracted using a standard CNN and the corresponding point cloud feature produced by the PointNet sub-network. Its job is to combine these features and to output a 3D bounding box for the target object. Below we propose two fusion network formulations, a vanilla \textit{global} fusion network, and a novel \textit{dense} fusion network.
\vparagraph{Global fusion network} As shown in Fig.~\ref{fig:model}C, the global fusion network processes the image and point cloud features and directly regresses the 3D locations of the eight corners of the target bounding box. We experimented with a number of fusion functions and found that a  concatenation of the two vectors, followed by applying a number of fully connected layers, results in optimal performance. The loss function with the global fusion network is then:
\begin{equation}
    L = \sum_i \mathrm{smoothL1}(\mathbf{x}^*_i, \mathbf{x}_i) + L_{\mathrm{stn}},
\end{equation}
where $\mathbf{x}^*_i$ are the ground-truth box corners, $\mathbf{x}_i$ are the predicted corner locations and $L_{stn}$ is the spatial transformation regularization loss introduced in ~\cite{qi2016pointnet} to enforce the orthogonality of the learned spatial transform matrix. 

A major drawback of the global fusion network is that the variance of the regression target $\mathbf{x}^*_i$ is directly dependent on the particular scenario. For autonomous driving, the system may be expected to detect objects from 1m to over 100m. This variance places a burden on the network and results in sub-optimal performance. To address this, we turn to the well-studied problem of 2D object detection for inspiration. Instead of directly regressing the 2D box, a common solution is to generate object proposals by using sliding windows~\cite{dalal2005histograms} or by predicting the box displacement relative to spatial anchors~\cite{redmon2016you,erhan2014scalable,huang2015densebox,ren2015faster}. These ideas motivate our \textit{dense fusion} network, which is described below.

\vparagraph{Dense fusion network} The main idea behind this model is to use the input 3D points as dense spatial anchors. Instead of directly regressing the absolute locations of the 3D box corners, for each input 3D point we predict the \textit{spatial offsets} from that point to the corner locations of a nearby box. As a result, the network becomes agnostic to the spatial extent of a scene. The model architecture is illustrated in Fig.~\ref{fig:model}C. We use a variant of PointNet that outputs point-wise features. For each point, these are concatenated with the global PointNet feature and the image feature resulting in an $n \times 3136$ input tensor. The dense fusion network processes this input using several layers and outputs a 3D bounding box prediction along with a score for each point. At test time, the prediction that has the highest score is selected to be the final prediction. Concretely, the loss function of the dense fusion network is:
\begin{equation}
    L = \frac{1}{N} \sum_i \mathrm{smoothL1} (\mathbf{x^i}^*_{\mathrm{offset}},\mathbf{x^i}_{\mathrm{offset}}) + L_{\mathrm{score}} + L_{\mathrm{stn}},
\end{equation}
where $N$ is the number of the input points, $\mathbf{x^i}^*_{\mathrm{offset}}$ is the offset between the ground truth box corner locations and the $i$-th input point, and $\mathbf{x^i}_{\mathrm{offset}}$ contains the predicted offsets. $L_{\mathrm{score}}$ is the score function loss, which we explain in depth in the next subsection. 
\vparagraph{3D box parameterization} We parameterize a 3D box by its 8 corners since: (1) The representation is employed in the current state-of-the-art methods~\cite{velofcn,mv3d}, which facilitates fair comparison. (2) It generalizes any 3D shapes with N reference points, and it works well with our spatial anchor scheme: we can predict the spatial offsets instead of the absolute locations of the corners.

\subsection{Dense Fusion Prediction Scoring}
\label{sec:score}
The goal of the $L_{\mathrm{score}}$ function is to focus the network on learning spatial offsets from points that are close to the target box. We propose two scoring functions: a \textit{supervised} scoring function that directly trains the network to predict if a point is inside the target bounding box and an \textit{unsupervised} scoring function that lets network to choose the point that would result in the optimal prediction. 
\vparagraph{Supervised scoring} The supervised scoring loss trains the network to predict if a point is inside the target box. Let's denote the offset regression loss for point $i$ as $L^i_{\mathrm{offset}}$, and the binary classification loss of the $i$-th point as $L^i_{\mathrm{score}}$. Then we have:
\begin{equation}
    L = \frac{1}{N}\sum_i (L^i_{\mathrm{offset}} \cdot m_i + L^i_{\mathrm{score}}) + L_{\mathrm{stn}},
\end{equation}
where $m_i \in \{0, 1\}$ indicates whether the $i$-th point is in the target bounding box and $L_{score}$ is a cross-entropy loss that penalizes incorrect predictions of whether a given point is inside the box. As defined, this supervised scoring function focuses the network on learning to predict the spatial offsets from points that are inside the target bounding box. However, this formulation might not give the optimal result, as the point most confidently inside the box may not be the point with the best prediction. 
\vparagraph{Unsupervised scoring}
The goal of \textit{unsupervised} scoring is to let the network learn directly which points are likely to give the best hypothesis, whether they are most confidently inside the object box or not. We need to train the network to assign high confidence to the point that is likely to produce a good prediction. The formulation includes two competing loss terms: we prefer high confidences $c_i$ for all points, however, corner prediction errors are scored proportional to this confidence. Let's define $L^i_{\mathrm{offset}}$ to be the corner offset regression loss for point $i$. Then the loss becomes: 
\begin{equation}
    L = \frac{1}{N}\sum (L^i_{\mathrm{offset}}\cdot c_i - w \cdot \log(c_i)) + L_{\mathrm{stn}}, 
\end{equation}
where $w$ is the weight factor between the two terms. Above, the second term encodes a logarithmic bonus for increasing $c_i$ confidences. We empirically find the best $w$ and use $w=0.1$ in all of our experiments.

\begin{figure*}[!t]
\centering
\includegraphics[width=\linewidth]{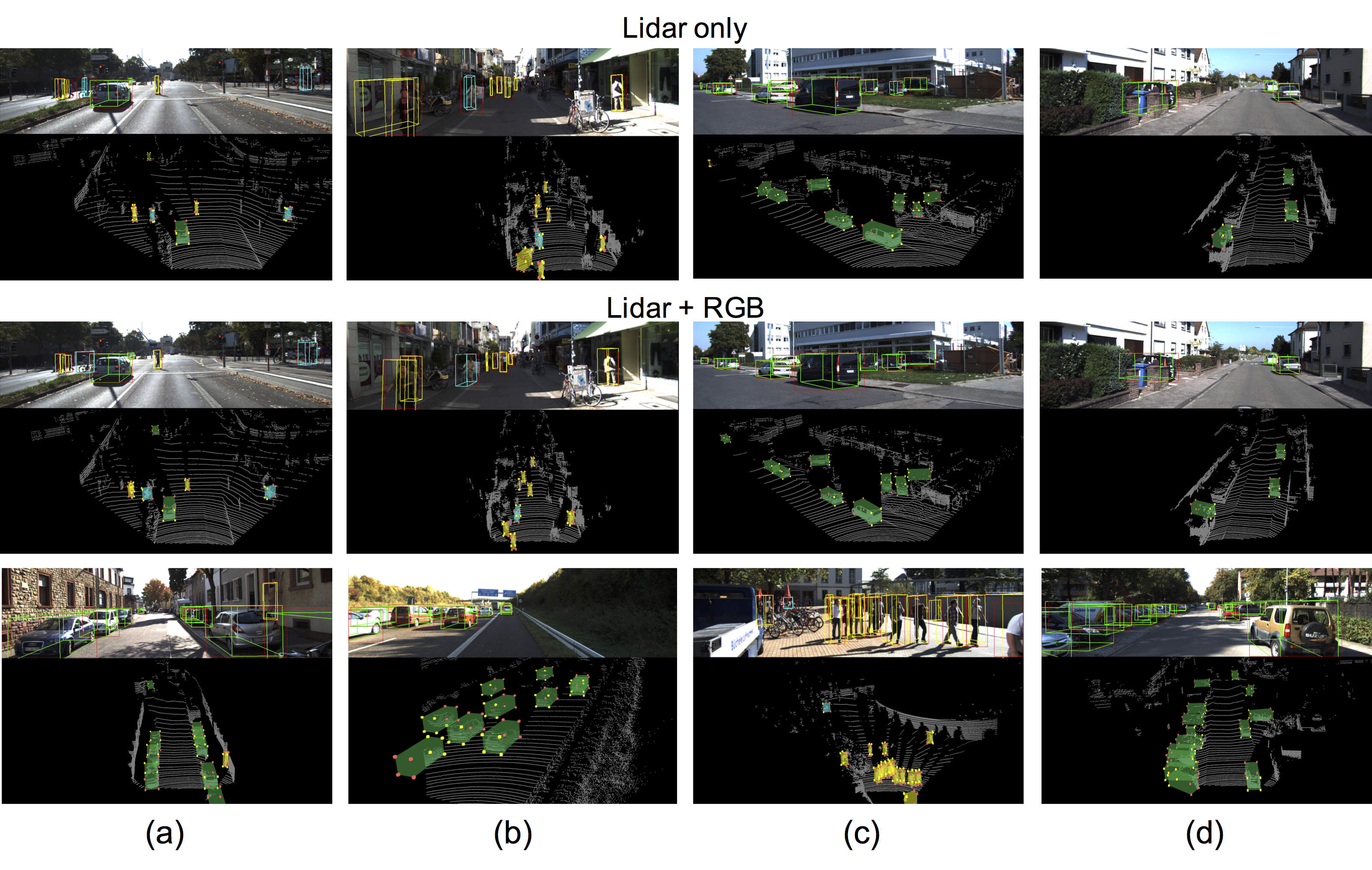}
\caption{Qualitative results on the KITTI dataset. Detection results are shown in transparent boxes in 3D and wireframe boxes in images. 3D box corners are colored to indicate direction: red is front and yellow is back. Input 2D detection boxes are shown in red. The top two rows compare our \textit{final} lidar + rgb model with a lidar only model ~\textit{dense-no-im}. The bottom row shows more results from the ~\textit{final} model. Detections with score $>0.5$ are visualized.}
\label{fig:kitti-qualitative}
\end{figure*}

\section{Experiments}

We focus on answering two questions: 1) does PointFusion perform well on different sensor configurations and environments compared to models that hold dataset or sensor-specific assumptions, and 2) do the dense prediction architectures perform better than architectures that directly regress the spatial locations. To answer 1), we compare our model against the state of the art on two distinctive datasets, the KITTI dataset~\cite{kitti} and the SUN-RGBD dataset~\cite{song2015sun}. To answer 2), we conduct ablation studies for the model variations described in Sec.~\ref{sec:model}.

\subsection{Datasets}
\paragraph{KITTI} The KITTI dataset~\cite{kitti} contains both 2D and 3D annotations of cars, pedestrians, and cyclists in urban driving scenarios. The sensor configuration includes a wide-angle camera and a Velodyne HDL-64E LiDAR. The official training set contains 7481 images. We follow~\cite{mv3d} and split the dataset into training and validation sets, each containing around half of the entire set. We report model performance on the validation set for all three
object categories.  

\paragraph{SUN-RGBD} The SUN-RGBD dataset~\cite{song2015sun} focuses on indoor environments, in which as many as 700 object categories are labeled. The dataset is collected via different types of RGB-D cameras with varying resolutions. The training and testing sets contain 5285 and 5050 images, respectively. We report model performance on the testing set.

We follow the KITTI training and evaluation setup with one exception. Because SUN-RGBD does not have a direct mapping between the 2D and 3D object annotations, for each 3D object annotation, we project the 8 corners of the 3D box to the image plane and use the minimum enclosing 2D bounding box as training data for the 2D object detector and our models. We report 3D detection performance of our models on the same 10 object categories as in~\cite{ren2016three,lahoud20172d}. Because these 10 object categories contain relatively large objects, we also show detection results on the 19 categories from ~\cite{song2016deep} to show our model's performance on objects of all sizes. We use the same set of hyper-parameters in both KITTI and SUN-RGBD.

\subsection{Metrics} We use the 3D object detection average precision metric  ($AP_{3D}$) in our evaluation. A predicted 3D box is a true positive if its \textit{3D intersection-over-union} ratio (3D IoU) with a ground truth box is over a threshold. We compute a per-class precision-recall curve and use the area under the curve as the AP measure. 

We use the official evaluation protocol for the KITTI dataset, i.e., the 3D IoU thresholds are \texttt{0.7, 0.5, 0.5} for \texttt{Car, Cyclist, Pedestrian} respectively. Following~\cite{song2015sun,ren2016three,lahoud20172d}, we use a 3D IoU threshold \texttt{0.25} for all classes in SUN-RGBD. 

\subsection{Implementation Details} 
\label{sec:implementation}
\noindent\textbf{Architecture} We use a ResNet-50 pretrained on ImageNet~\cite{imagenet} for processing the input image crop. The output feature vector is produced by the final residual block (block-4) and averaged across the feature map locations. We use the original implementation of PointNet with all batch normalization layers removed. For the 2D object detector, we use an off-the-shelf Faster-RCNN~\cite{ren2015faster} implementation~\cite{tfdetapi} pretrained on MS-COCO~\cite{mscoco} and fine-tuned on the datasets used in the experiments. We use the same set of hyper-parameters and architectures in all of our experiments.

\noindent\textbf{Training and evaluation} During training, we randomly resize and shift the ground truth 2D bounding boxes by 10\% along their $x$ and $y$ dimensions. These boxes are used as the input crops for our models. At evaluation time, we use the output of the trained 2D detector. For each input 2D box, we crop and resize the image to $224 \times 224$ and randomly sample a maximum of 400 input 3D points in both training and evaluation. At evaluation time, we apply PointFusion to the top 300 2D detector boxes for each image. The 3D detection score is computed by multiplying the 2D detection score and the predicted 3D bounding box scores. 

\subsection{Architectures}
We compare 6 model variants to showcase the effectiveness of our design choices.
\begin{itemize}[leftmargin=*]
    \setlength\itemsep{1pt}
    \item \textit{final} uses our dense prediction architecture and the unsupervised scoring function as described in Sec.~\ref{sec:score}.
    \item \textit{dense} implements the dense prediction architecture with a supervised scoring function as described in Sec.~\ref{sec:score}.
    \item \textit{dense-no-im} is the same as \textit{dense} but takes only the point cloud as input.
    \item \textit{global} is a baseline model that directly regresses the 8 corner locations of a 3D box, as shown in Fig.~\ref{fig:model}D.
    \item \textit{global-no-im} is the same as the \textit{global} but takes only the point cloud as input.
    \item \textit{rgb-d} replaces the PointNet component with a generic CNN, which takes a depth image as input. We use it as an example of a homogeneous architecture baseline. \footnote{We have experimented with a number of such architectures and found that achieving a reasonable performance requires  non-trivial effort. Here we present the model that achieves the best performance. Implementation details of the model are included in the supplementary material.}
\end{itemize}

\subsection{Evaluation on KITTI}
\paragraph{Overview} Table~\ref{table:kitti-car} shows a comprehensive comparison of models that are trained and evaluated only with the \texttt{car} category on the KITTI validation set, including all baselines and the state of the art methods 
3DOP~\cite{3dop} (stereo), VeloFCN~\cite{velofcn} (LiDAR), and MV3D~\cite{mv3d} (LiDAR + rgb).
Among our variants, \textit{final} achieves the best performance, while the homogeneous CNN architecture \textit{rgb-d} has the worst performance, which underscores the effectiveness of our heterogeneous model design. 

\begin{table}
\centering
\caption{AP$_{3D}$ results for the \texttt{car} category on the KITTI dataset.  Models are trained on \texttt{car} examples only, with the exception of \textit{Ours-final (all-class)}, which is trained on all 3 classes.}
\label{table:kitti-car}
\begin{tabular}{|l|l|ccc|}
\hline
method & input & easy     & mod.   & hard     \\ \hline
3DOP\cite{3dop}         & Stereo                                      & 12.63    & 9.49       & 7.59           \\
VeloFCN\cite{velofcn}   & 3D                                       & 15.20    & 13.66      & 15.98       \\
MV3D~\cite{mv3d}        & 3D + rgb                                 & 71.29    & 62.68      & \textbf{56.56}  \\
rgb-d                    & 3D + rgb                                & 7.43     & 6.13       & 4.39      \\
Ours-global-no-im       & 3D                                       & 28.83    & 21.59      & 17.33   \\
Ours-global             & 3D + rgb                                 & 43.29    & 37.66      & 32.23    \\
Ours-dense-no-im        & 3D                                       & 62.13    & 42.31      & 34.41    \\
Ours-dense              & 3D + rgb                                 & 71.53    & 59.46      & 49.41  \\
Ours-final              & 3D + rgb                                 & 74.71    & 61.24      & 50.55  \\
Ours-final (all-class)  & 3D + rgb                                 & \textbf{77.92}    & \textbf{63.00}      & 53.27 \\

\hline
\end{tabular}
\end{table}

\noindent\textbf{Compare with MV3D~\cite{mv3d}} The \textit{final} model also outperforms the state of the art method MV3D~\cite{mv3d} on the easy category (3\% more in $AP_{3D}$), and has a similar performance on the moderate category (1.5\% less in $AP_{3D}$). When we train a single model using all 3 KITTI categories \textit{final (all-class)}, we roughly get a 3\% further increase, achieving
a 6\% gain over MV3D on the easy examples and a 0.5\% gain on the moderate ones. This shows that our model learns a generic 3D representation that can be shared across categories. Still, MV3D outperforms our models on the hard examples, which are objects that are significantly occluded, by a considerable margin (6\% and 3\% $AP_{3D}$ for the two models mentioned). We believe that the gap with MV3D for occluded objects is due to two factors: 1) MV3D uses a bird's eye view detector for cars, which is less susceptible to occlusion than our front-view setup. It also uses custom-designed features for car detection that should generalize better with few training examples 2) MV3D is an end-to-end system, which allows one component to potentially correct errors in another. Turning our approach into a fully end-to-end system may help close this gap further. 
Unlike MV3D, our general and simple method achieves excellent results on \texttt{pedestrian} and \texttt{cyclist}, which are state of the art by a large margin (see Table~\ref{table:kitti-all}).

\noindent\textbf{Global vs. dense} The \textit{dense} architecture has a clear advantage over the \textit{global} architecture as shown in Table~\ref{table:kitti-car}: \textit{dense} and \textit{dense-no-im} outperforms \textit{global} and \textit{global-no-im}, respectively, by large margins. This shows the effectiveness of using input points as spatial anchors. 

\noindent\textbf{Supervised vs unsupervised scores} In Sec.~\ref{sec:score}, we introduce a \textit{supervised} and an \textit{unsupervised} scoring function formulation. Table~\ref{table:kitti-car} and Table~\ref{table:kitti-all} show that the unsupervised scoring function performs a bit better for our car-only and all-category models. These results support our hypothesis that a point confidently inside the object is not always the point that will give the best prediction. It is better to rely on a self-learned scoring function for the specific task than on a hand-picked proxy objective.

\begin{table}[t!]
\centering
\caption{AP$_{3D}$ results for models trained on all KITTI classes.}
\label{table:kitti-all}
\begin{tabular}{c|cccc}
\hline
category                    & model & easy  & moderate & hard  \\ \hline
\multirow{3}{*}{car}        & Ours-no-im & 55.68 & 39.85    & 33.71  \\
                            & Ours-dense & 74.77 & 61.42    & 51.88 \\
                            & Ours-final & \textbf{77.92} & \textbf{63.00}    & \textbf{53.27} \\ \hline
\multirow{3}{*}{pedestrian} & Ours-no-im & 4.61  & 2.58     & 3.58 \\
                            & Ours-dense & 31.91 & 26.82    & 22.59 \\
                            & Ours-final & \textbf{33.36} & \textbf{28.04}    & \textbf{23.38} \\ \hline
\multirow{3}{*}{cyclist}    & Ours-no-im & 3.07  & 2.58     &  2.14 \\
                            & Ours-dense & 47.21 & 28.87    & 26.99 \\
                            & Ours-final & \textbf{49.34} & \textbf{29.42}    & 26.98 \\ \hline
\end{tabular}
\end{table}

\begin{figure}[]
\centering
\includegraphics[width=0.8\linewidth]{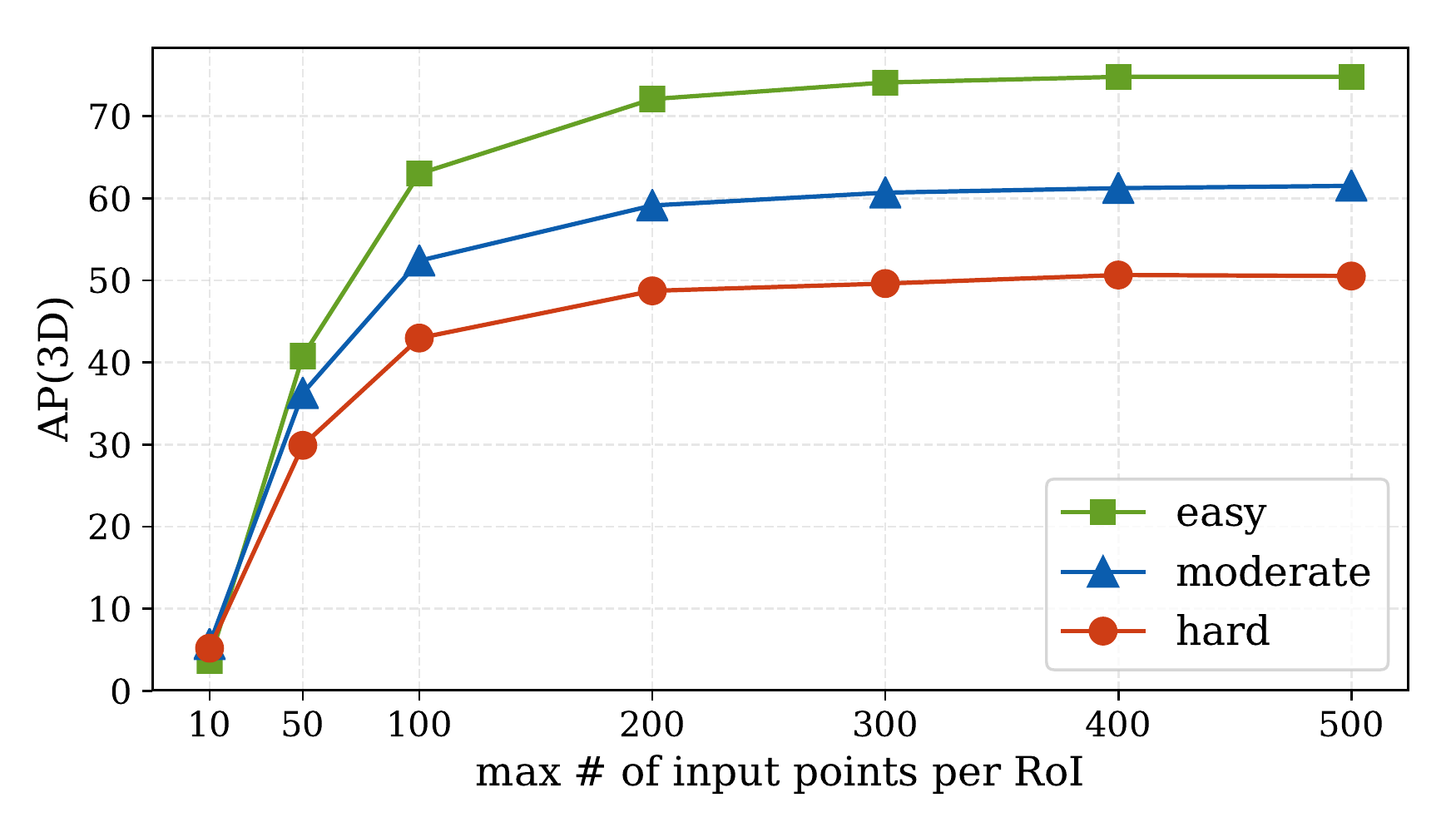}
\caption{Ablation experiment: 3D detection performances ($AP_{3D}$) given maximum number of input points per RoI.}
\label{fig:num-point}
\end{figure}

\noindent\textbf{Effect of fusion} Both car-only and all-category evaluation results show that fusing lidar and image information always yields significant gains over lidar-only architectures, but the gains vary across classes. Table~\ref{table:kitti-all} shows that the largest gains are for \texttt{pedestrian} (3\% to 47\% in $AP_{3D}$ for easy examples) and for \texttt{cyclist} (5\% to 32\%). Objects in these categories are smaller and have fewer lidar points, so they benefit the most from high-resolution camera data. Although sparse lidar points often suffice in determining the spatial location of an object, image appearance features are still helpful in estimating the object dimensions and orientation. This effect is analyzed qualitatively below.

\begin{table*}[t!]
\centering
\caption{3D detection results on the SUN-RGBD test set using the 3D Average Precision metrics with 0.25 IoU threshold. Our model achieves results that are comparable to the state-of-the-art models while achieving much faster speed.}
\small
\label{table:sun}
\begin{tabular}{c|cccccccccc|cc}
\hline
 Method                            & bathtub & bed   & bkshelf   & chair  & desk  & dresser & n. stand    & sofa  & table & toilet & mAP   & runtime  \\ \hline

DSS~\cite{song2016deep}            & 44.2          & 78.8          & 11.9      & 61.2   & 20.5  & 6.4     & 15.4        & 53.5  & 50.3  & 78.9   & 42.1  & 19.6s    \\
COG~\cite{ren2016three}            & \textbf{58.26}& 63.67         & 31.80     & \textbf{62.17}  & \textbf{45.19} & 15.47   & 27.36       & 51.02 & \textbf{51.29} & 70.07  & 47.63 & 10-30m \\
Lahoud \etal.~\cite{lahoud20172d}  & 43.45         & 64.48         & 31.40     & 48.27  & 27.93 & \textbf{25.92}   & \textbf{41.92}       & 50.39 & 37.02 & 80.4   & 45.12 & 4.2s    \\
rgbd                               & 36.78         & 60.44         & 20.48     & 46.11  & 12.71 & 14.35   & 30.19       & 46.11 & 24.80 & 81.79  & 38.17 & 0.9s  \\
Ours-dense-no-im                   & 28.68         & 66.55         & 22.43     & 50.70  & 13.86 & 14.20   & 25.69       & 49.74 & 23.63 & 83.35  & 39.24 & 0.4s  \\
Ours-final                         & 37.26         & \textbf{68.57}& \textbf{37.69}     & 55.09  & 17.16 & 23.95   & 32.33       & \textbf{53.83} & 31.03 & \textbf{83.80}  & 45.38 & 1.3s  \\ \hline
\end{tabular}
\end{table*}

\noindent\textbf{Qualitative Results} Fig.~\ref{fig:kitti-qualitative} showcases some sample predictions from the lidar-only architecture \textit{dense-no-im} and our \textit{final} model. We observe that the fusion model is better at estimating the dimension and orientation of objects than the lidar-only model. In column (a), one can see that the fusion model is able to determine the correct orientation and spatial extents of the cyclists and the pedestrians whereas the lidar-only model often outputs inaccurate boxes. Similar trends can also be observed in (b). In (c) and (d), we note that although the lidar-only model correctly determines the dimensions of the cars, it fails to predict the correct orientations of the cars that are occluded or distant. The third row of Fig.~\ref{fig:kitti-qualitative} shows more complex scenarios. (a) shows that our model correctly detects a person on a ladder. (b) shows a complex highway driving scene. (c) and (d) show that our model may occasionally fail in extremely cluttered scenes.

\noindent\textbf{Number of input points}
Finally, we conduct a study on the effect of limiting the number of input points at test time. Given a \textit{final} model trained with at most 400 points per crop, we vary the maximum number of input points per RoI and evaluate how the 3D detection performance changes. As shown in~\ref{fig:num-point}, the performance stays constant at 300-500 points and degrades rapidly below 200 points. This shows that our model needs a certain amount of points to perform well but is also robust against variations. 

\subsection{Evaluation on SUN-RGBD}
\textbf{Comparison with our baselines} As shown in Table~\ref{table:sun},  \textit{final} is our best model variant and  outperforms the \textit{rgb-d} baseline by 6\% mAP. This is a much smaller gap than in the KITTI dataset, which shows that the CNN performs well when it is given dense depth information (rgb-d cameras provide a depth measurement for every rgb image pixel). Furthermore, \textit{rgb-d} performs roughly on-par with our lidar-only model, which demonstrates the effectiveness of our PointNet subcomponent and the dense architecture. 

\textbf{Comparison with other methods} We compare our model with three approaches from the current state of the art. Deep Sliding Shapes (DSS)~\cite{song2016deep} generates 3D regions using a proposal network and then processes them using a 3D convolutional network, which is prohibitively slow. Our model outperforms DSS by 3\% mAP while being 15 times faster. Clouds of Oriented Gradients (COG) by Ren \etal ~\cite{ren2016three} exploits the scene layout information and performs exhaustive 3D bounding box search, which makes it run in the tens of minutes. In contrast, PointFusion only uses the 3D points that project to a 2D detection box and still outperforms COG on 6 out of 10 categories, while approaching its overall mAP performance. PointFusion also compares favorably to the method of Lahoud \etal ~\cite{lahoud20172d}, which uses a multi-stage pipeline to perform detection, orientation regression and object refinement using object relation information. Our method is simpler and does not make environment-specific assumptions, yet it obtains a marginally better mAP while being 3 times faster. Note that for simplicity, our evaluation protocol passes all 300 2D detector proposals for each image to PointFusion. Since our 2D detector takes only 0.2s per frame, we can easily get sub-second evaluation times simply by discarding detection boxes with scores below a threshold, with minimal performance losses. 

\begin{figure}[]
\centering
\includegraphics[width=\linewidth]{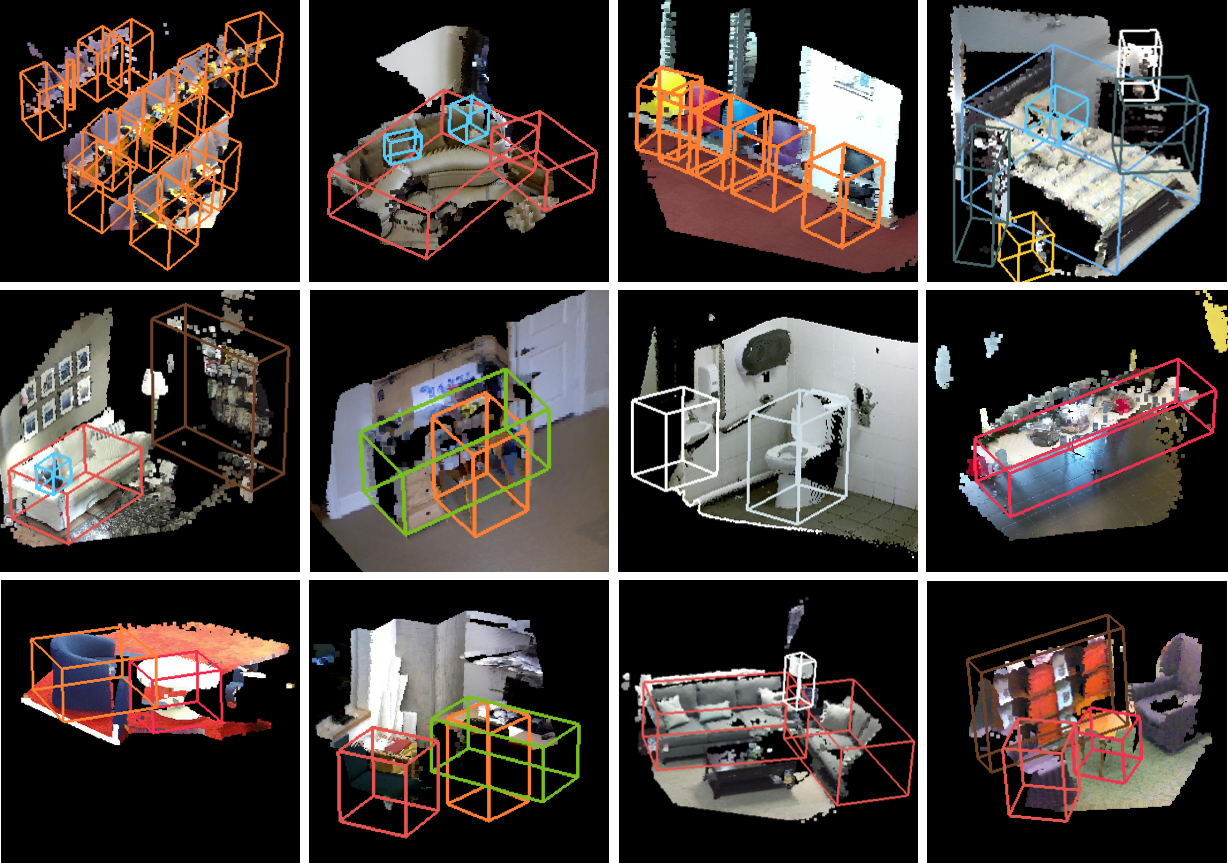}
\caption{Sample 3D detection results from our ~\textit{final} model on the SUN-RGBD test set. Our modle is able to detect objects of variable scales, orientations, and even objects on top of other objects. Detections with score $> 0.7$ are visualized.}
\label{fig:sun-qualitative}
\end{figure}

\textbf{Qualitative results} Fig.~\ref{fig:sun-qualitative} shows some sample detection results from the \textit{final} model on 19 object categories. Our model is able to detect objects of various scales, orientations, and positions. Note that because our model does not use a top-down view representation, it is able to detect objects that are on top of other objects, e.g., pillows on top of a bed. Failure modes include errors caused by objects which are only partially visible in the image or from cascading errors from the 2D detector. 

\section{Conclusions and Future Work}
We present the PointFusion network, which accurately estimates 3D object bounding boxes from image and point cloud information. Our model makes two main contributions. First, we process the inputs using heterogeneous network architectures. The raw point cloud data is directly handled using a PointNet model, which avoids lossy input preprocessing such as quantization or projection. Second, we introduce a novel dense fusion network, which combines the image and point cloud representations. It predicts multiple 3D box hypotheses relative to the input 3D points, which serve as spatial anchors, and automatically learns to select the best hypothesis. We show that with the same architecture and hyper-parameters, our method is able to perform on par or better than methods that hold dataset and sensor-specific assumptions on two drastically different datasets. Promising directions of future work include combining the 2D detector and the PointFusion network into a single end-to-end 3D detector, as well as extending our model with a temporal component to perform joint detection and tracking in video and point cloud streams.

{\small
\bibliographystyle{ieee}
\bibliography{egbib}
}

\section{Supplementary}
\subsection{3D Localization AP in KITTI} In addition to the $AP_{3D}$ metrics, we also report results on a 3D localization $AP_{loc}$ metrics just for reference.  A predicted 3D box is a true positive if its 2D top-down view box has an IoU with a ground truth box is greater than a threshold.  We compute a per-class precision-recall curve and use the area under the curve as the AP measure. 
We use the official evaluation protocol for the KITTI dataset, i.e., the 3D IoU thresholds are \texttt{0.7, 0.5, 0.5} for \texttt{Car, Cyclist, Pedestrian} respectively. Table~\ref{table:kitti-car-supp} shows the results on models that are trained on \texttt{Car} only, with the exception of \textit{final (all-class)}, which is trained on all categories, and Table~\ref{table:kitti-all-supp} shows the results of models that are trained on all categories. 

\subsection{The \textit{rgbd} baseline} In the experiment section, we show that the \textit{rgbd} baseline model performs the worst on the KITTI dataset. We observe that most of the predicted boxes have less than 0.5 IoU with ground truth boxes due to the errors in the predicted depth. The performance gap is reduced in the SUN-RGBD dataset due to the availability of denser depth map. However, it is non-trivial to achieve such performance using a CNN-based architecture. Here we describe the \textit{rgbd} baseline in detail. 

\subsubsection{Input representation} The \textit{rgbd} baseline is a CNN architecture that takes as input a 5-channel tensor. The first three channels is the input RGB image. The fourth channel is the depth channel. For KITTI, we obtain the depth channel by projecting the lidar point cloud on to the image plane, and assigning zeros to the pixels that have no depth values. For SUN-RGBD, we use the depth image. We normalize the depth measurement by the maximum depth range value. The fifth channel is a depth measurement binary mask: 1 indicates that the corresponding pixel in the depth channel has a depth value. This is to add extra information to help the model to distinguish between no measurements and small measurements. Empirically we find this extra channel useful.

\subsubsection{Learning Objective} We found that training the model to predict the 3D corner locations is ineffective due to the highly non-linear mapping and lack of image grounding. Hence we regress the box corner pixel locations and the depth of the corners and then use the camera geometry to recover the full 3D box. A similar approach has been applied in ~\cite{mousavian20163d}. The pixel regression target is normalized between 0 and 1 by the dimensions of the input 2D box. For the depth objective, we found that directly regressing the depth value is difficult especially for the KITTI dataset, in which the target objects have large location variance. Instead, we employed a multi-hypothesis method: we discretize the depth objective into overlapping bins and train the network to predict which bin contains the center of the target box. The network is also trained to predict the residual depth of each corner to the center of the predicted depth bin. At test time, the corner depth values can be recovered by adding up the center depth of the predicted bin and the predicted residual depth of each corner. Intuitively, this method lets the network to have a coarse-to-fine estimate of the depth, alleviating the large variance in the depth objective.

\begin{table*}[!t]
\centering
\caption{3D detection results ($AP_{3D}$) of the \texttt{car} category on the KITTI dataset. We compare against a number of state-of-the-art models. }
\label{table:kitti-car-supp}
\begin{tabular}{|l|l|ccc|ccc|}
\hline
\multirow{2}{*}{method} & \multicolumn{1}{c|}{\multirow{2}{*}{input}} & \multicolumn{3}{c|}{AP$_{3D}$} & \multicolumn{3}{c|}{AP$_{loc}$}  \\ \cline{3-8} 
                        & \multicolumn{1}{c|}{}                       & easy     & moderate   & hard    & easy     & moderate   & hard       \\ \hline
3DOP\cite{3dop}         & Stereo                                      & 12.63    & 9.49       & 7.59    & 12.63    & 9.49       & 7.59          \\
VeloFCN\cite{velofcn}   & lidar                                       & 15.20    & 13.66      & 15.98   & 40.14    & 32.08      & 30.47       \\
MV3D~\cite{mv3d}        & lidar + rgb                                 & 71.29    & 62.68      & \textbf{56.56}   & 86.55    & 78.10      & \textbf{76.67}     \\
rgb-d                    & lidar + rgb                                & 7.43     & 6.13       & 4.39    & 10.40    & 7.77       & 5.83      \\
Ours-global-no-im       & lidar                                       & 28.83    & 21.59      & 17.33   & 40.01    & 32.48      & 27.34   \\
Ours-global             & lidar + rgb                                 & 43.29    & 37.66      & 32.23   & 54.49    & 48.01      & 39.19    \\
Ours-dense-no-im        & lidar                                       & 62.13    & 42.31      & 34.41   & 76.31    & 57.61      & 47.66    \\
Ours-dense              & lidar + rgb                                 & 71.53    & 59.46      & 49.41   & 83.61    & 73.12      & 62.54    \\
Ours-final              & lidar + rgb                                 & 74.71    & 61.24      & 50.55   & \textbf{90.53}    & \textbf{78.13}      & 65.41   \\
Ours-final (all-class)  & lidar + rgb                                 & \textbf{77.92}    & \textbf{63.00}      & 53.27   & 87.45    & 76.13      & 65.32   \\

\hline
\end{tabular}
\end{table*}

\begin{table*}[!t]
\centering
\caption{3D detection ($AP_{3D}$) results of all categories on the KITTI dataset.}
\label{table:kitti-all-supp}
\begin{tabular}{|l|l|lll|lll|}
\hline
\multirow{2}{*}{category}   & \multirow{2}{*}{model} & \multicolumn{3}{c|}{$AP_{3D}$} & \multicolumn{3}{c|}{$AP_{loc}$} \\ \cline{3-8} 
                            &                        & easy    & moderate   & hard    & easy     & moderate   & hard    \\ \hline
\multirow{3}{*}{Car}        & global-no-im           & 55.68   & 39.85      & 33.71   & 72.35    & 55.42      & 47.41   \\
                            & dense                  & 74.77   & 61.42      & 51.88   & 47.21    & 28.87      & 26.99    \\
                            & final                  & 77.92   & 63.00         & 53.27   & 49.34    & 29.42      & 26.98   \\ \hline
\multirow{3}{*}{Pedestrian} & global-no-im           & 3.07    & 2.58       & 3.58    & 3.07     & 2.58       & 2.14    \\
                            & dense                  & 31.96   & 26.82      & 22.59   & 37.08    & 31.71      & 27.04   \\
                            & final                  & 33.36   & 28.04      & 23.38   & 37.91    & 32.35      & 27.35   \\ \hline
\multirow{3}{*}{Cyclist}    & global-no-im           & 4.61    & 2.58       & 3.58    & 6.93     & 3.76       & 4.7     \\
                            & dense                  & 47.21   & 28.87      & 26.99   & 51.45    & 31.65      & 29.59  \\
                            & final                  & 49.34   & 29.42      & 26.98   & 54.02    & 32.77      & 30.19   \\ \hline
\end{tabular}
\end{table*}

\end{document}